\documentclass{article}

\usepackage{PRIMEarxiv}

\usepackage[utf8]{inputenc} 
\usepackage[T1]{fontenc}    
\usepackage{hyperref}       
\usepackage{url}            
\usepackage{booktabs}       
\usepackage{amsfonts}       
\usepackage{nicefrac}       
\usepackage{microtype}      
\usepackage{lipsum}
\usepackage{fancyhdr}       
\usepackage{graphicx}       
\graphicspath{{media/}}     
\usepackage{enumerate}
\usepackage{mathtools}
\usepackage{float}
\usepackage{pifont}

\usepackage{amsopn, amssymb, amscd, amsmath, amsthm}

\usepackage[ruled]{algorithm2e}

\usepackage{tikz-cd}

\usepackage{physics}

\usepackage{bbm}

\DeclareMathOperator{\lime}{LIME}
\DeclareMathOperator{\argmin}{argmin}
\DeclareMathOperator{\ReLU}{ReLU}
\DeclareMathOperator{\diag}{diag}
\DeclareMathOperator{\IG}{IG}
\DeclareMathOperator{\Rob}{Rob}
\DeclareMathOperator{\LLE}{LLE}
\DeclareMathOperator{\AS}{AS}

\theoremstyle{plain}
\newtheorem{theorem}{Theorem}[section]

\theoremstyle{definition}
\newtheorem{definition}[theorem]{Definition}

\theoremstyle{remark}

\providecommand{\dd}{\mathrm{d}}
\providecommand{\exp}{\mathrm{exp}}

\providecommand{\tr}{\mathrm{tr}}

\title{Probabilistic Lipschitzness and The Stable Rank for Comparing Explanation Models}

\author{
  Lachlan Simpson$^{1}$, Kyle Millar$^{2}$, Adriel Cheng$^{1,2}$, Cheng-Chew Lim$^{1}$, and Hong Gunn Chew$^{1}$\\
  $^{1}$School of Electrical and Mechanical Engineering, The University of Adelaide, Australia \\
  $^{2}$Information Sciences Division, Defence Science \& Technology Group, Australia \\
  \texttt{\{lachlan.simpson, honggunn.chew, adriel.cheng, cheng.lim\}@adelaide.edu.au}\\
  \texttt{\{kyle.millar1, adriel.cheng\}}@defence.gov.au
}

\begin{document}
\maketitle

\begin{abstract}
Explainability models are now prevalent within machine learning to address the black-box nature of neural networks. The question now is which  explainability model is most effective. Probabilistic Lipschitzness has demonstrated that the smoothness of a neural network is fundamentally linked to the quality of post hoc explanations. In this work, we prove theoretical lower bounds on the probabilistic Lipschitzness of Integrated Gradients, LIME and SmoothGrad. We propose a novel metric using probabilistic Lipschitzness, normalised astuteness, to compare the robustness of explainability models. Further, we prove a link between the local Lipschitz constant of a neural network and its stable rank. We then demonstrate that the stable rank of a neural network provides a heuristic for the robustness of explainability models.
\end{abstract}


\section{Introduction}

Neural networks have become state-of-the-art solutions to a wide array of tasks from computer vision \cite{yolo}, natural language processing \cite{Liu_2023}, cyber-security \cite{9737249}, to fundamental problems in biology \cite{alphafold}. An unsolved problem in machine learning is how neural networks achieve their high performance \cite{Sejnowski_2020}. The limited understanding of how neural networks learn is known as the black-box problem \cite{zednik2019solving}. The black-box problem becomes critical as neural networks have become prevalent within safety critical systems such as self driving cars. The black-box problem is of increasing concern when a neural network fails. Furthermore, neural networks have been shown to be susceptible to adversarial attacks in applications that include facial recognition \cite{dong2019efficient} and self driving cars \cite{8578273}. Neural network exploits were achieved by observing that a small change in the input resulted in a large change in the output \cite{szegedy2014intriguing}.

Post hoc explainability models are methods for explaining the features that influence the output of a neural network in a human understandable format \cite{ieeeXAIsurvey}. Post hoc explainability is a step towards addressing the black-box problem \cite{zednik2019solving}. Post hoc explainability can be categorised in two forms: perturbation and gradient based \cite{agarwal2021unification, ieeeGraphXAI}. Perturbation techniques such as LIME \cite{ribeiro2016why} and SHAP \cite{lundberg2017unified} locally approximate complex neural networks to simple models (e.g. decision trees and linear models). Gradient methods such as Integrated Gradients \cite{pmlr-v70-sundararajan17a} and SmoothGrad \cite{smilkov2017smoothgrad} provide an explanation of a neural network through analysing its gradient. The gradient of a neural network provides information of the features that produce the largest change in output.

Explainability models are used to elucidate neural network predictions in essential domains, including cancer screening \cite{ ieeeXAIsurvey,LAMY201942} and cyber-security \cite{Zhang_2022}. To ensure trust-worthy explanations, an explainability model must yield consistent explanations for close data points \cite{agarwal2021unification,khan2023analyzing}. 
The magnitude of the Lipschitz constant of an explainability model quantifies the similarity of explanations for close data points.  \cite{agarwal2021unification, khan2023analyzing,yeh2019infidelity,Montavon_2018}.

The Lipschitz constant is a useful property to analyse model robustness. However, the Lipschitz constant is provably hard to calculate \cite{scaman2019lipschitz, jordan2021exactly}. Recent experimental work has shown that a low rank approximation of the embedding matrix of a neural network, the stable rank, is related to the Lipschitz constant of a classifier \cite{ramasinghe2022periodicity}. The stable rank has then been proposed as a heuristic measure of the magnitude of the Lipschitz constant due to its ease of calculation in polynomial time. In this work, we make the first connection between the stable rank and explainability robustness.

The analysis of the Lipschitz constant of explainability models SHAP and LIME offer valuable insights into the local robustness of the explanations of both models \cite{alvarezmelis2018robustness}. Three metrics had been proposed for explainability robustness: local Lipschitz estimate \cite{alvarezmelis2018robustness}, average sensitivity \cite{yeh2019infidelity}, and astuteness \cite{khan2023analyzing}. Robustness metrics should be bounded, independent of other explainability models under analysis and provide a single measure of robustness over all points in the dataset. Local Lipschitz estimate and average sensitivity are unbounded and evaluated point-wise, and are unsuitable as single quantifiable measure of robustness. 

Probabilistic Lipschitzness provides a probability of local robustness\cite{10.1007/978-3-030-65474-0_13} for classifiers. Astuteness extends probabilistic Lipschitzness to explainability models. Astuteness thereby provides a probabilistic measure of the local robustness of explainability models.  Astuteness is bounded and is not evaluated point-wise. Astuteness therefore provides a single measure of robustness for the explainability model. The limitation of astuteness is a dependency on the choice of Lipschitz constant. Different Lipschitz constants will result in different measures of robustness. In this work, we propose \textit{normalised astuteness}, a variant of astuteness without a dependency on an ad hoc choice of Lipschitz constant.

The contributions of this work are as follows:
\begin{enumerate}
    \item We prove lower bounds for the astuteness of three prevalent explainability models, Integrated Gradients, LIME, and SmoothGrad.
    
    \item We propose normalised astuteness as a metric to compare explainability model robustness. We compare normalised astuteness with two robustness metrics, local Lipschitz estimate and average sensitivity on well-known machine learning datasets.
    
    \item We prove a lower bound relationship between the Lipschitz constant of a classifier and its stable rank. We establish a relationship between probabilistic Lipschitzness and stable rank. We use this relationship to demonstrate the efficacy of stable rank as a heuristic measure of explainability model robustness.

\end{enumerate}

The remainder of this paper is structured as follows: Section 2 introduces related work. In Section 3, we define probabilistic Lipschitzness and astuteness for classifiers and explainability models. In Section 4, we prove lower bound theoretical guarantees for three explainability models: Integrated Gradients, LIME, and SmoothGrad. In Section 5, we introduce astuteness as a metric for measuring the robustness of explainability models. We  provide a comparative analysis of our metric against two metrics proposed in literature; local Lipschitz estimate and average sensitivity. We then validate the use of our metric to assess the robustness of three explainability models across three well-know datasets. In Section 6, we prove a lower bound estimate of the Lipschitz constant which depends on the stable rank. This bound is progress towards establishing a theoretical relationship between the Lipschitz constant and the stable rank. We then demonstrate that the stable rank can be used as a heuristic measure of the robustness of explainability models. We conclude in Section 7 with a discussion of future work. 

\subsection{Notation and Assumptions}
We suppose our points live in an open and bounded subset $ x \in U \subseteq \mathbb{R}^n$ and sampled from a distribution $\mathcal{D}$. We denote a trained neural network as $f \colon U \to \mathbb{R}^m$ and suppose $f$ is sufficiently smooth (i.e. $f \in \mathcal{C}^{\infty}$). An explanation model is denoted as a function $\psi \colon U \to \mathbb{R}^n$ where $\psi(x)_{i}$ is the attribution of the \textit{i}-th feature. We measure the distance between two points $x,x'\sim D$ with the standard p-norm denoted as $\|-\|_{p}$. Lastly, we consider two points $x,y \sim D$ as \textit{close} if for a given $r \geq 0$,\; $\|x-y\|_{p} \leq r$.

\section{Related Work}
\textbf{Explainer robustness and robustness metrics:} 
Robustness of explainability models has varying definitions within the literature \cite{khan2023analyzing}. 
We adopt the definition of robustness as \textit{the stability of explanations for points in a local neighbourhood} \cite{alvarezmelis2018robustness}.
The magnitude of the Lipschitz constant of an explainability model quantifies the stability of local explanations. Robustness is a crucial property of explainability models for two reasons. First, a robust explainability model can locally replace a complex neural network \cite{alvarezmelis2018robustness}. Second, non-robust explainability models provide conflicting information for users seeking to comprehend a model's behaviour \cite{arrieta2019explainable}.

Robustness of explainability models has been shown to depend on the local smoothness of the neural network \cite{khan2023analyzing}. The extent that a neural network is locally smooth is formulated probabilistically with \textit{probabilistic Lipschitzness} \cite{10.1007/978-3-030-65474-0_13}. \textit{Astutenss} has extended probabilistic Lipschitzness to explainability models \cite{khan2023analyzing}. Astuteness provides a probability that an explainability method is locally robust.

Theoretical guarantees on the astuteness of explainability models CXPLAIN, RISE and SHAP are provided in \cite{khan2023analyzing}. Robustness (as defined in \cite{alvarezmelis2018robustness}) of Continuous LIME (C-LIME) and SmoothGrad have been shown to depend on the maximum gradient of the neural network \cite{agarwal2021unification}. For a neural network with a closed interval domain, Integrated Gradients may fail to be Lipschitz \cite{lundstrom2022rigorous}. However, if the gradient of the neural network is Lipschitz then Integrated Gradients is Lipschitz \cite{lundstrom2022rigorous}. We however assume that our points live in an open and bounded subset $ U \subseteq \mathbb{R}^n$ as in \cite{jordan2021exactly}. With the assumption of $U$ being open and bounded, we prove lower bound guarantees on the astuteness of Integrated Gradients, LIME, and SmoothGrad. The theoretical bounds presented herein strengthen the hypothesis in \cite{khan2023analyzing} that robust explainability models are determined by the smoothness of the underlying neural network. 

In literature, metrics for assessing explainability model robustness are lacking. Three metrics, however,  have been proposed, local Lipschitz estimate, average sensitivity, and astuteness.
Local Lipschitz estimate approximates the maximum Lipschitz constant of a point in a local neighbourhood. Local Lipschitz estimate provides a worst-case estimate of local robustness. A similar metric to local Lipschitz estimate is average sensitivity. Average sensitivity takes the average estimated local Lipschitz constant. Both metrics are unbounded and are evaluated point-wise, making them unsuitable as a single quantifiable measure of robustness. Astuteness has been proposed as a metric for robustness. The limitation of astuteness is a dependency on a choice of Lipschitz constant. Different choices in the Lipschitz constant result in different measures of robustness. In this work, we propose a variant use of astuteness without a dependency on the choice of Lipschitz constant.   

\textbf{Stable rank and the Lipschitz constant:}\:
The Lipschitz constant is a well studied property of neural networks \cite{10.1007/978-3-030-65474-0_13,ieeeLip,gouk2020regularisation}. Learning a low Lipschitz constant has been shown to improve generalisation. Gouk et al. \cite{gouk2020regularisation} showed that a neural network will yield a greater performance if it is constrained to have a low Lipschitz constant. Furthermore, Khan et al. \cite{khan2023analyzing} demonstrate that enforcing a low Lipschitz constant of a classifier improves explainability model robustness.  

Computing the Lipschitz constant with $\ell_{2}$ norm is NP-Hard and provably hard to calculate with $\ell_{1}$ and $\ell_{\infty}$ norms, scaling by a factor of the input dimension \cite{jordan2021exactly}. Whilst the Lipschitz constant of a neural network is hard to compute \cite{ieeeLip}, the stable rank which is an approximation of the rank of a matrix is not. The link between the stable rank and Lipschitz constant of neural networks has been demonstrated in a number of domains. Generalisation bounds on neural networks depend on both the stable rank of the weight matrix and Lipschitz constant \cite{NIPS2017_b22b257a, pmlr-v40-Neyshabur15}. The algorithm, \textit{stable rank normalisation} demonstrates that simultaneously optimising both the stable rank and Lipschitz constant yields improved generalisation and classification performance \cite{sanyal2020stable}. In signal reconstruction, the magnitude of the stable rank of the hidden-layer representations is related to magnitude of the Lipschitz constant \cite{ramasinghe2022periodicity}. In \cite{ramasinghe2022periodicity}, the empirical observation is made that a high stable rank results in a high Lipschitz constant and vice-versa.

To the best of the authors knowledge this is the first theoretical and experimentally validated connection between the stable rank and robustness of explainability models.

\section{Probabilistic Lipschitzness and Astuteness}
In this section we define Probabilistic Lipschitzness for classifiers and Astuteness for explainability models. 
\begin{definition}{\cite{10.1007/978-3-030-65474-0_13} Probabilistic Lipschitzness:}
        Given a probability and distance threshold $ 0 \leq \alpha \leq 1, r\geq 0$, a function $f \colon U \to \mathbb{R}$ is probabilistically Lipschitz with constant $L \geq 0$ if
        \begin{equation*}
            \mathbb{P}_{x,x'\sim \mathcal{D}}(d_p(f(x),f(x')) \leq Ld_{p}(x,x') \mid d_{p}(x,x') \leq r) \geq 1-\alpha.
        \end{equation*}
        \end{definition}
        Probabilistic Lipschitzness naturally extends to explainability models.
        \begin{definition}{\cite{khan2023analyzing} Explainer astuteness:} The astuteness of an explainer $\psi$ over $\mathcal{D}$, denoted as $A_{r,\lambda}(\psi, \mathcal{D})$ is the probability that $\forall x,x' \sim \mathcal{D}$ such that $d_{p}(x,x') \leq r$, the explanations generated by $\psi$ are at most $\lambda \:d_{p}(-)$ away from each other, where $\lambda \geq 0$.
        \begin{equation*}
            A_{r,\lambda}(\psi,\mathcal{D}) = \mathbb{P}_{x,x'\sim \mathcal{D}}\left(d_p(\psi(x),\psi(x')) \leq \lambda d_{p}(x,x') \mid d_{p}(x,x') \leq r \right) \geq 1-\alpha.
        \end{equation*}
        \end{definition}

        The key idea from \cite{khan2023analyzing} is that the local smoothness of a neural network and the astuteness of an explainability model are linked. In practice this means that a smoother neural network results in locally robust post hoc explanations.

\section{Astuteness Theoretical Guarantee}
In this section, we prove theoretical bounds on the astuteness of LIME, Integrated Gradients, and SmoothGrad. The theoretical results presented herein, further establish the claim in \cite{khan2023analyzing} that the smoothness of a neural network dictates the robustness of an explainability model.
\subsection{Integrated Gradients}
For a neural network $f \colon U \to \mathbb{R}^{n}$, we fix a point $x' \in U$, called the base-point. Given a point to explain $x \in U$ we define a straight line path $\gamma_x \colon [0,1] \to U$ as $\gamma_{x}(\alpha) = (1-\alpha)x' +\alpha x$. Here there is an implicit assumption that the function $f$ and its gradient $\nabla f$ is defined along the path $\gamma_{x}$. We define integrated gradients as follows.
\begin{definition} \cite{pmlr-v70-sundararajan17a}
Given a base-point $x' \in U$ and an explanation point $x \in U$, we define integrated gradients as
\begin{equation*}
    \IG(x,x') = (x-x') \odot \int_{\gamma_{x}} \nabla f(\gamma_{x}(\alpha)) \dd \alpha,
\end{equation*}
where $\odot$ is the Hadamard product. For a single feature attribution we have
\begin{equation*}
    \IG(x,x')_{i} = (x_{i}-x'_{i})\int_{\gamma_{x_{i}}} \partial_{i}f(\gamma_{x_{i}}(\alpha)) \dd \alpha.
\end{equation*}
    
\end{definition}
        \begin{theorem}
        Suppose $f$ is probabilistic $L$-Lipschitz with probability $\geq 1 - \alpha$. Take two points $x,y \sim \mathcal{D}$ such that $d_{p}(x,y) \leq r$ and a base-point $x' \sim \mathcal{D}$. Then for integrated gradients we have $A_{r, \lambda}(\IG) \geq 1- \alpha$, where, 

        \begin{equation*}
            \lambda = 3L\sqrt{n}\frac{\sup_{x\neq y }\|x-y\|}{\inf_{x \neq y}\|x-y\|}.
        \end{equation*}
        \end{theorem}

        \begin{proof}
Suppose $f$ is L-Lipschitz. Then by Rademacher's theorem \cite{Federer_2005}, $f$ is differentiable a.e. and $\|\nabla f\| \leq L$ \cite{scaman2019lipschitz}. Take $x,y \sim D$ with pairwise distance at most $r$.
        \begin{align*}
            \| \IG(x,x') - \IG(y,x') \| &= \| (x - x') \odot \int_{0}^{1} \nabla f(\gamma_{x})\dd\alpha -(y - x')\odot\int_{0}^{1} 
             \nabla f(\gamma_{y})\dd\alpha  \|\\
             &=\| (x - y) \odot \int_{0}^{1} \nabla f(\gamma_{x})\dd\alpha +(y - x') \odot \int_{0}^{1} (\nabla f(\gamma_{x}) - \nabla f(\gamma_{y}))\dd\alpha \| \\
              &\leq \sqrt{n} \| x - y \|\int_{0}^{1} \|\nabla f(\gamma_{x})\|\dd\alpha +\sqrt{n} \|y - x'\|\int_{0}^{1} \| \nabla f(\gamma_{x}) - \nabla f(\gamma_{y})\|\dd\alpha \\
            &\leq \sqrt{n}\| x - y \|\int_{0}^{1} \|\nabla f(\gamma_{x})\|\dd\alpha + \|y - x'\|\sqrt{n}\int_{0}^{1} \|\nabla f(\gamma_{y})\|+\|\nabla f(\gamma_{x})\|\dd\alpha \\
            &\leq L\sqrt{n}\|x - y\| + 2L\sqrt{n} \|y - x'\|\\
            &\leq 3L\sqrt{n} \sup_{x \neq y}\|x - y\|.
        \end{align*}
        Now we require some $\lambda \geq 0 $ such that 
        \begin{equation*}
            3L\sqrt{n} \sup_{x \neq y}\|x- y\|\leq \lambda \|x - y\|.
        \end{equation*}
        It is clear that $\lambda$ as defined above satisfies the inequality. It then follows that
        \begin{equation*}
          \mathbb{P}\left( \left\|\IG(x,x') ,\IG(y,x')\right\| \leq \lambda \| x- y \| \mid  \|x- y\| \leq r , \| f(x) - f(y)\| \leq L \|x - y\| \right) = 1.  
        \end{equation*}
        
        Now we are interested in the conditional probability 
        \begin{equation*}
        \mathbb{P}(\|\IG(x,x') - \IG(y,x')\| \leq \lambda \|x-y\| \mid \|x- y\| \leq r).
        \end{equation*}
       
        Since $f$ is probabilistic Lipschitz we have
         \begin{equation*}
         \mathbb{P}( \|f(x)- f(y)\| \leq L \|x-y\| \mid \|x- y\| \leq r) \geq 1- \alpha.    
         \end{equation*}

        Marginalising over the event $\|f(x) -  f(y)\| \leq L\|x-y\|$, we have 
        \begin{equation*}
            \mathbb{P}(\| \IG(x,x') - \IG(y,x') \| \leq \lambda\|x - y\| \mid \|x - y\| \leq r) \geq 1 - \alpha.
        \end{equation*}
        Therefore, $A_{r,\lambda}(\IG) \geq 1-\alpha$.
        \end{proof}

\subsection{LIME}
LIME is an explainability model from the class of perturbation model explainers. LIME approximates a simple model (e.g. linear models, decision trees) to a neighbourhood of a point $x \in U$ to locally describe the behaviour of a complex decision boundary. We use the formulation of LIME as defined in \cite{agarwal2021unification}.
\begin{definition}
Given a point $x \in U$ define $S_{x} \subseteq U$ as points in a local neighbourhood of $x$. The points in $S_{x}$ are weighted by a distance metric $\pi \colon U \cross U \to \mathbb{R}^{\geq 0}$. The distance metric $\pi$ is usually the Gaussian kernel
\begin{equation*}
 \pi(x,y) = \exp(\frac{-\|x-y\|^2}{\sigma^2}).   
\end{equation*}
For a class of models $G$ and complexity measure $\Omega$, LIME solves the following optimisation problem
\begin{equation*}
\lime(x) = \argmin_{g \in G} \mathcal{L}(f,g,\pi, S_{x}) + \Omega(g),
\end{equation*}
where $\mathcal{L}$ is the weighted square difference loss function defined as
\begin{equation*}
 \mathcal{L}(f,g,\pi, S_{x}) =\frac{1}{|S_{x}|} \sum_{a \in S_{x}}\pi(x,a)\|f(a) - g(a)\|^2.
\end{equation*}
We take $G$ to be linear models and $\Omega$ as the number of non-zero coefficients.
\end{definition}
\begin{theorem}
Suppose f is probabilistic L-Lipschitz with probability $\geq 1-\alpha$ and take $x,y \sim D$ such that $d_{p}(x,y) \leq r$. Then for $\lime$ we have $A_{r, \lambda}(\lime) \geq 1-\alpha$, where, 
\begin{equation*}
    \lambda = L + \frac{C}{\inf_{x \neq y}\|x-y\|},
\end{equation*}
with
\begin{equation*}
 C = 2\sqrt{2|D|+L^2r^2}.  
\end{equation*}
\end{theorem}
\begin{proof}
Suppose $g \in G$ is a solution to the LIME optimisation problem. For ease of notation we neglect the $\Omega(g)$ term. The goal is to find some $\lambda > 0$ such that for all $x,y \sim \mathcal{D}$
\begin{equation*}
 \|g(x)-g(y)\| \leq \lambda\|x-y\|.
\end{equation*} We note by the triangle inequality that
\begin{align*}
    \|g(x)-g(y)\| &\leq \|f(x)-g(x)\|+\|f(x)-g(y)\|\\
    &\leq \|f(x)-g(x)\| + \|f(x)-f(y)\|+\|f(y)-g(y)\|\\
    &\leq \|f(x)-g(x)\|+\|f(y)-g(y)\|+L\|x-y\|.
\end{align*} We will find bounds for each term on the RHS of the above inequality.

Since 
            \begin{equation*}
                \sum_{a \in S_{x}}\pi(x,a)\|f(a) - g(a)\|^2
            \end{equation*}
            is a positive finite sum, it contains the term $\pi(x,x)\|f(x)-g(x)\|^2$ and $\pi(x,x) = 1$. So we have
            \begin{equation*}
                \|f(x) - g(x)\|^2 \leq \sum_{a \in S_{x}}\pi(x,a)\|f(a) - g(a)\|^2.
            \end{equation*}

    Since $g \in G$ is the linear model which minimises the above sum, for any other linear model $h \in G$ we have the following inequality.         
    \begin{equation*}
    \|f(x) - g(x)\|^2 \leq \sum_{a \in S_{x}}\pi(x,a)\|f(a) - g(a)\|^2
    \leq \sum_{a \in S_{x}}\pi(x,a)\|f(a) - h(a)\|^2.
    \end{equation*}
    Now choose the linear model $h \in G$ as the constant function
    \begin{equation*}
    h(a) \equiv f(y) \text{ for all } a \in S_{x}.
    \end{equation*}
    Note that we can rewrite the sum on the RHS as
            \begin{equation*}
            \sum_{a \in S_{x}}\pi(x,a)\|f(a) - h(a)\|^2  = \sum_{a \in S_{x}\backslash \{x\}}\pi(x,a)\|f(a) - f(y)\|^2 + \|f(x) - f(y)\|^2.
            \end{equation*}

            Recall that $f$ is assumed to be $L$-Lipschitz and therefore
            \begin{equation*}
             \|f(x) - g(x)\|^{2} \leq \sum_{a \in S_{x}\backslash \{x\}}\pi(x,a)\|f(a) - f(y)\|^2 + L^2\|x-y\|^2.
             \end{equation*}
             A similar argument shows
             \begin{equation*}
             \|f(y) - g(y)\|^{2} \leq \sum_{a \in S_{y}\backslash \{y\}}\pi(y,a)\|f(a) - f(x)\|^2 + L^2\|x-y\|^2.
             \end{equation*}

             For brevity, let us define
             \begin{equation*}
                 C_1 = \sqrt{\sum_{a \in S_{y}\backslash \{y\}}\pi(y,a)\|f(a) - f(x)\|^2 + L^2\|x-y\|^2}
                 \end{equation*}
                 and
                 \begin{equation*}
                 C_2 = \sqrt{\sum_{a \in S_{x}\backslash \{x\}}\pi(x,a)\|f(a) - f(y)\|^2 + L^2\|x-y\|^2}.
                \end{equation*}
            Then,
             \begin{equation*}
              \|g(x)-g(y)\| \leq C_1 + C_2 + L\|x-y\|.
             \end{equation*}
         We now require some constant $\lambda > 0$ such that
        \begin{equation*}
            \|g(x) - g(y)\| \leq \lambda \|x-y\|.
        \end{equation*}
        Note that the Gaussian kernel achieves it's maximum at $x = y$. Thus, for all $x,y \in U$
        \begin{equation*}
        \pi(x,y) \leq 1. 
        \end{equation*}
        We can further bound the constant $C_1$.
        \begin{align*}
            C_1^2 &\leq \sum_{a \in S_{y}\backslash \{y\}}\|f(a)-f(y)\|^2 + L^2\|x-y\|^2\\
            &\leq \sum_{a \in S_{y}\backslash \{y\}} \sup_{x \neq y}\|f(x)-f(y)\|^2+L^2\|x-y\|^2\\
            &= (|S_{y}|-1)\sup_{x \neq y}\|f(x)-f(y)\|^2+L^2\sup_{x \neq y}\|x-y\|^2\\
            &\leq |D|\sup_{x \neq y}\|f(x)-f(y)\|^2+L^2\sup_{x \neq y}\|x-y\|^2\\
            &\leq |D|\sup_{x \neq y}\|f(x)-f(y)\|^2+L^2r^2.
        \end{align*}
        The difference between $f(x)$ and $f(y)$ is maximised when $x,y$ are different classes, i.e. $x_{c} \neq y_{c}$. Thus,
        \begin{equation*}
            \sup_{x_{c}\neq y_{c}}\|f(x)-f(y)\|^{2} = 2.
        \end{equation*}
        Hence,
        \begin{equation*}
            C_{1}^{2} \leq 2|D| + L^2r^2.
        \end{equation*}
        Likewise, $C_2$ has the same upper bound. Let us define
        \begin{equation*}
            C = 2\sqrt{2|D|+L^2r^2}.
        \end{equation*}
        So,
         \begin{equation*}
              \|g(x)-g(y)\| \leq C + L\|x-y\|.
         \end{equation*}
         Then for
         \begin{equation*}
             \lambda = L+\frac{C}{\inf_{x\neq y}\|x-y\|},
         \end{equation*}
         we have
         \begin{equation*}
             \|g(x)-g(y)\| \leq \lambda\|x-y\|.
         \end{equation*}
         Now we are interested in the conditional probability 
        \begin{equation*}
        \mathbb{P}(\|\lime(x) - \lime(y)\| \leq \lambda \|x-y\| \mid \|x- y\| \leq r). 
        \end{equation*}
       
        From above 
        \begin{equation*}
          \mathbb{P}( \|\lime(x) - \lime(y)\| \leq \lambda \| x- y \| \mid  \|x- y\| \leq r , \| f(x) - f(y)\| \leq L \|x - y\| ) = 1.  
        \end{equation*}
         and by $f$ being probabilistic Lipschitz we have
         \begin{equation*}
         \mathbb{P}( \|f(x)- f(y)\| \leq L \|x-y\| \mid \|x- y\| \leq r) \geq 1- \alpha.    
         \end{equation*}

        Marginalising over the event $\|f(x) -  f(y)\| \leq L\|x-y\|$, we have 
        \begin{equation*}
            \mathbb{P}(\| \lime(x) - \lime(y) \| \leq \lambda\|x - y\| \mid \|x - y\| \leq r) \geq 1 - \alpha.
        \end{equation*}
        Therefore, $A_{r,\lambda}(\lime) \geq 1-\alpha$.
\end{proof}
\subsection{SmoothGrad}
\begin{definition}{\cite{smilkov2017smoothgrad}}
Let $x \in U$ and $S_{x} \subseteq U$ be a finite neighbourhood of $x$. Suppose $f \colon U \to \mathbb{R}^{m}$ is a trained neural network. We define SmoothGrad as
\begin{equation*}
    SG_{S_{x}}^{f}(x) = \frac{1}{|S_{x}|}\sum_{a \in S_{x}} \nabla f(a).
\end{equation*}
\end{definition}

\begin{theorem}
Let $x,y \sim D$ such that $d_{p}(x,y) \leq r$ and $ 0 \leq \alpha  \leq 1$. Suppose $f$ is probabilistic $L$-Lipschitz with probability $\geq 1 -\alpha$. Suppose $x$ and $y$ have finite sample neighbourhoods of equal size (i.e. $|S_{x}| = |S_{y}|$). Then $A_{r, 
\lambda}(\text{SG}) \geq 1 - \alpha$,
where
\begin{equation*}
    \lambda = \frac{2L}{\inf_{x\neq y} \|x - y\|}.
\end{equation*}
\end{theorem}

\begin{proof}
    Take $x,y \sim D$ with $d_{p}(x,y) \leq r$ and let $|S_{x}| = |S_{y}| = K \in \mathbb{N}$.
    \begin{align*}
        \|SG(x)-SG(y)\| &= \Big \|\frac{1}{|S_{x}|}\sum_{a \in S_{x}} \nabla f(a) -\frac{1}{|S_{y}|}\sum_{b \in S_{y}} \nabla f(b) \Big \|\\
        &\leq \frac{1}{K}( \sum_{a \in S_{x}} \| \nabla f(a)\| + \sum_{b \in S_{y}}\| \nabla f(b)) \|)\\
        &\leq\frac{1}{K}(K\cdot L + K \cdot L)\\
        &\leq \frac{1}{K}( 2KL)\\
        &= 2L.
    \end{align*}
    Now choose
    \begin{equation*}
        \lambda = \frac{2L}{\inf_{x \neq y} \|x-y\|},
    \end{equation*}
    then
    \begin{equation*}
        \|SG(x)-SG(y)\| \leq \lambda \| x -y \|.
    \end{equation*}
    Now we are interested in the conditional probability 
        \begin{equation*}
        \mathbb{P}(\|SG(x) - SG(y)\| \leq \lambda \|x-y\| \mid \|x- y\| \leq r).
        \end{equation*}
        From above 
        \begin{equation*}
          \mathbb{P}( \|SG(x)-SG(y)\| \leq \lambda \| x- y \| \mid  \|x- y\| \leq r , \| f(x) - f(y)\| \leq L \|x - y\| ) = 1.  
        \end{equation*}
         and by $f$ being probabilistic Lipschitz we have
         \begin{equation*}
         \mathbb{P}( \|f(x)- f(y)\| \leq L \|x-y\| \mid \|x- y\| \leq r) \geq 1- \alpha.    
         \end{equation*}

        Marginalising over the event $\|f(x) -  f(y)\| \leq L\|x-y\|$, we have 
        \begin{equation*}
            \mathbb{P}(\| SG(x) - SG(y) \| \leq \lambda\|x - y\| \mid \|x - y\| \leq r) \geq 1 - \alpha.
        \end{equation*}
        Therefore, $A_{r,\lambda}(SG) \geq 1-\alpha$.
\end{proof}

\section{Normalised Astuteness as a Metric for Comparing Explanation Models}
In this section, a variant of astuteness is proposed as a robustness metric for explainability models. First, we discuss three robustness metrics within literature: Local Lipschitz estimate (LLE), average sensitivity (AS), and astuteness. Second, we define four criteria a robustness metric must meet. Last, we compare normalised astuteness with LLE and AS.
\subsection{Robustness metrics and criteria for an adequate metric }
Robustness is defined as the sensitivity of a model to small changes in points. Given two sufficiently close points $x,y \in U$ the difference of a model output $\|f(x)- f(y)\|$ is measured relative to the difference in input $\|x-y\|$. Model robustness is quantified via the magnitude of the Lipschitz constant, defined as
\begin{equation*}
    \|f(x)-f(y)\|\leq L \|x-y\|.
\end{equation*}

A model with high Lipschitz constant may be highly sensitive to changes in the input. Lipschitzness is usually defined globally. In practice, neural networks are unlikely to meet global Lipschitzness \cite{10.1007/978-3-030-65474-0_13}. For explainability, a global notion of smoothness does not provide adequate insight into robustness. Distant points need not have similar explanations, Lipschitzness is therefore restricted to a local measure. Three metrics have been proposed to measure the local robustness of explanation models: LLE, AS and astuteness (Definition 3.2).

\begin{definition}{ \cite{alvarezmelis2018robustness} Local Lipschitz estimate}
\begin{equation*}
    \LLE(\psi, x) = \max_{y \in N_{\varepsilon}(x)} \frac{\|\psi(x) - \psi(y)\|_{2}}{\|x-y\|_{2}}.
\end{equation*}
    
\end{definition}

\begin{definition} { \cite{yeh2019infidelity} Average Sensitivity}
\begin{equation*}
\AS(\phi, x) = \mathbb{E}_{y \in N_{\varepsilon}(x)} \frac{\| \psi(x) - \psi(y)\|_{2}}{\|x-y\|_{2}},
\end{equation*}
\end{definition}
where $N_{\varepsilon}(x) = \{y \in U \mid d(x,y) \leq \varepsilon\}$ is the $\varepsilon$-neighbourhood of $x$.

We define four criteria a robustness metric must meet:
\begin{enumerate}
    \item Boundedness:
    Boundedness ensures a robustness metric has an optimal and minimal measure.
    \item Model Independence:
    Model independence ensures that the metric is only dependent on the current model under analysis and not a collection of models. 
    \item Point Independence: Point independence ensures that a metric provides single quantifiable measure over all points rather than on singular points.
    \item Minimal hyper-parameters: Greater number of hyper-parameters adds complexity in the interpretation of the robustness measure. We require a metric with hyper-parameters that are as simple as possible. As robustness is defined locally, we take the minimal hyper-parameter to be the radius, $r$ which defines the maximal distance between two points.
\end{enumerate}

\subsection{The limitations of current robustness metrics}
LLE and AS have two limitations. First, both LLE and AS are evaluated pointwise. Whilst capturing information of local explainability robustness about a point, LLE and AS do not provide a single measure of robustness for the whole dataset. Second, both metrics are unbounded. Bommer et al. \cite{bommer2023finding} propose the following normalisation to address the unboundedness of LLE and AS. Suppose we have a set of $n$ explanation models $\{\psi_{i}\}_{i = 1}^{n}$. Let $\Rob(\psi_{i})$ be the evaluation of $\psi_{i}$ on LLE or AS.
Then one can normalise $\Rob(\psi_{m})$ as
\begin{equation*}
    \frac{\min_{i} \Rob(\psi_{i})}{\Rob(\psi_m)}.
\end{equation*}
Note that this normalisation depends on all explainability models under analysis. The set of explainability models $\{\phi_{i}\}$, therefore, influences the robustness measure of an explainability model. Robustness metrics should be bounded and independent of other models. LLE and AS with the normalisation of Bommer et al. \cite{bommer2023finding} therefore violates: boundedness, model, and point independence.

A further drawback of AS, LLE, and the normalisation of \cite{bommer2023finding} is the interpretation of the metrics. The most one can extract from these metrics is the magnitude of the local Lipschitz constant for each explanation model for a given point. Local explainability robustness about a point can then be inferred. One can aggregate LLE and AS over all points to reconcile the point-wise limitation of the metrics. The choice of aggregation, however, adds another layer required to interpret the result of the metric.

Astuteness as a robustness metric to compare explanation models is proposed in \cite{khan2023analyzing}. To define astuteness as a comparison metric, Khan et al. \cite{khan2023analyzing} fix a Lipschitz constant and compute the astuteness of each explanation model on a dataset. Astuteness is then used to rank each explanation model. The approach of \cite{khan2023analyzing} addresses two limitations of LLE and AS.  First, astuteness is bounded and not evaluated pointwise. Second, the interpretation of astuteness is intuitive and simple. Astuteness provides (for a given Lipschitz constant) the probability that close by points have similar explanations. The choice of Lipschitz constant then dictates the degree of robustness measured. 

The astuteness metric is dependent on a choice of Lipschitz constant which is not an obvious choice a priori. Different choices of Lipschtiz constant provide different measures of robustness.

\subsection{Normalised astuteness as a robustness metric}
In this section we introduce normalised astuteness. Normalised astuteness is a novel extension of astuteness which meets all criteria for a robustness metric as outlined in Section 5.1. Given an explainability model $\psi$ and an interval $[0, \lambda]$, we define the \textit{astuteness curve} as the graph by computing the astuteness $A_{r,x}(\psi)$ over the interval $x \in [0,\lambda]$. Below we list the steps for normalised astuteness:

\begin{enumerate}
    \item For each explainability model $\psi_{i}$, calculate the astuteness $A_{r,\lambda_{i}}(\psi_{i})$ and increase $\lambda_{i}$ until $A_{r, \lambda_{i}}(\psi_{i}) = 1$.
    \item Normalise the Lipschitz axis of each astuteness curve by $1/\lambda_{i}$ such that $A_{r, \lambda_{i}} = 1$.
    \item Take the AUC of each normalised astuteness curve.
\end{enumerate}

We consider a model robust if astuteness converges quickly for small Lipschitz constant. One can interpret this as given two close by points (of distance at most $r$) the probability that the points have similar explanations is high. The higher the AUC the smaller the Lipschitz constant for converging astuteness. Therefore a high AUC corresponds to a highly robust explainability model. In Table \ref{table:table_0} we compare normalised astuteness with LLE, AS, and astuteness. We see that only normalised astuteness satisfies all four criteria for a robustness metric.
\begin{table}[t]
\centering
\caption{A comparison of each robustness metric with the metric criteria defined in Section 5.1.}
\begin{tabular}[t]{llllll}
\toprule
 Metric Criteria& Normalised Astuteness & Astuteness & LLE & AS \\
\midrule
Bounded & Yes & No  & No & No \\
Model Independence  & Yes & Yes & Yes & Yes\\
Point Independence  &Yes& Yes & No & No \\
Minimal Hyper-parameters & Yes& No & Yes & Yes\\
\bottomrule
\label{table:table_0}
\end{tabular}
\end{table}%

\subsection{The efficacy of normalised astuteness as a robustness metric}
In this section, normalised astuteness is compared against LLE and AS with two explainability models, Integrated Gradients and SHAP. Integrated Gradients and SHAP are selected as example models from two types of explainability, gradient and perturbation based.

First, we visually analyse the explanations provided by Integrated Gradients and SHAP. Second, we look at LLE, AS and normalised astuteness on both explainability models. We see that all metrics align with our intuition of the robustness of Integrated Gradients and SHAP. Normalised astuteness, however, provides a single quantifiable measure of robustness whereas the other metrics do not. Third, we apply normalised astuteness to SHAP, LIME and Integrated Gradients on the XOR, Iris and MNIST datasets.

\subsubsection{A comparison of Integrated Gradients and SHAP explanations on the XOR problem}

We compare the normalised astuteness, LLE and AS of Integrated Gradients and SHAP on the XOR problem for a 2-layer MLP with ReLU activation. We take $r$ and $\varepsilon$ to be median pairwise distance as in \cite{khan2023analyzing}. In Figure \ref{fig:fig1}, we have the explanations of Integrated Gradients and SHAP on the XOR problem represented as vectors. The direction and magnitude of the explanation vector indicates which dimension the explanation technique has placed the highest attribution for a prediction. Note that the explanation vectors of SHAP and Integrated Gradients are not directly comparable in terms of explanation. The feature attribution of SHAP is the extent a feature moves a prediction away or towards the mean prediction. We see in Figure \ref{fig:fig1}, that for the blue class the explanation vectors move towards a positive prediction and the red class explanation vectors move towards a negative prediction. For Integrated Gradients, the explanation vectors are interpreted as the features with the highest impact on a prediction. We can see intuition behind explanations in Figure~\ref{fig:fig1}, in the lower left quadrant of the XOR problem where points closer to the decision boundary $X0 = 0$ have a higher importance in the $X0$ feature whereas points closer to $X1 = 0$ have a higher importance in the $X1$ feature.

\subsubsection{A comparison of normalised astuteness with LLE and AS}
\begin{figure}[t]
  \centering
  \includegraphics[scale = 0.6]{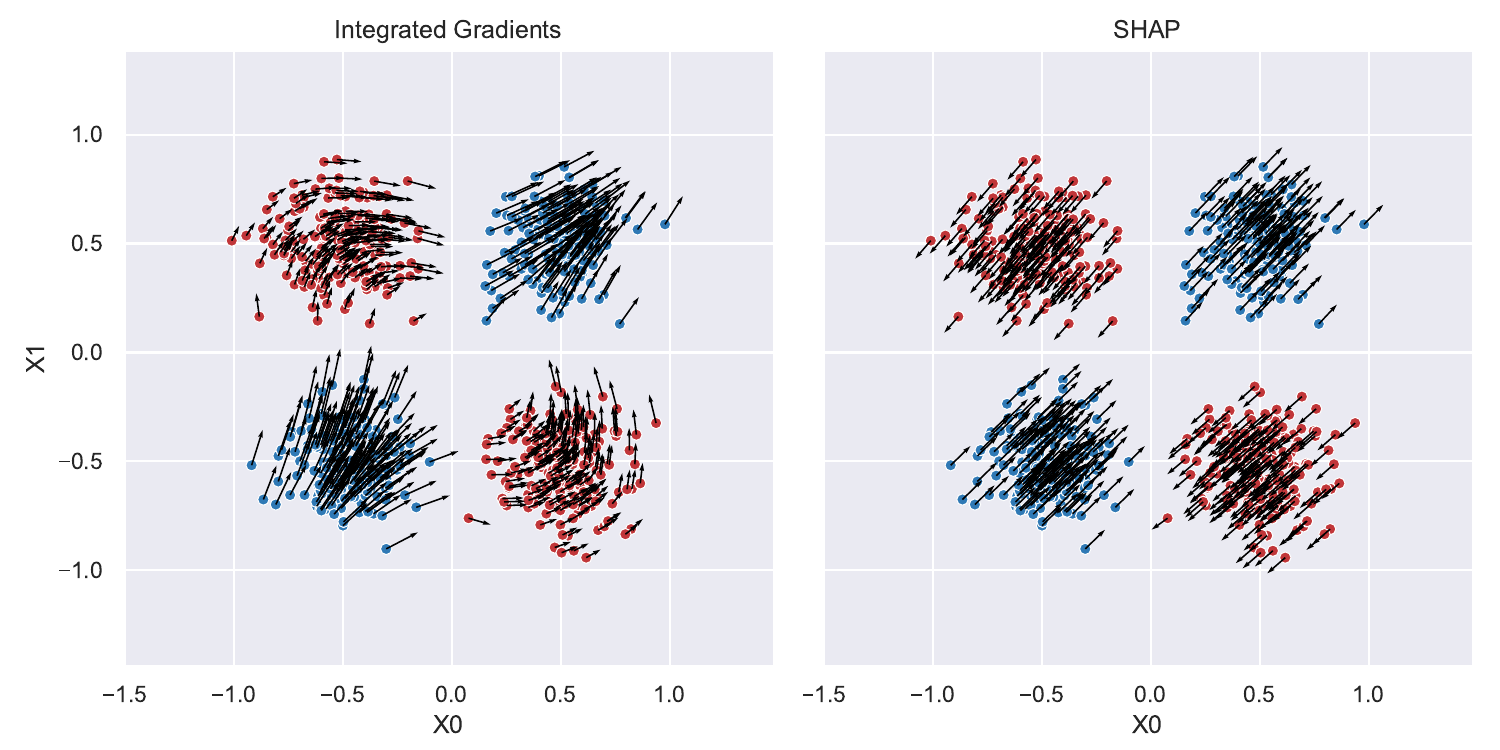}
  \caption{Integrated Gradients and SHAP explanation vectors on XOR problem. Direction and magnitude of explanation vectors indicate feature contribution to a prediction.}
  \label{fig:fig1}
\end{figure}
\begin{figure}[t]
  \centering
  \includegraphics[scale = 0.5]{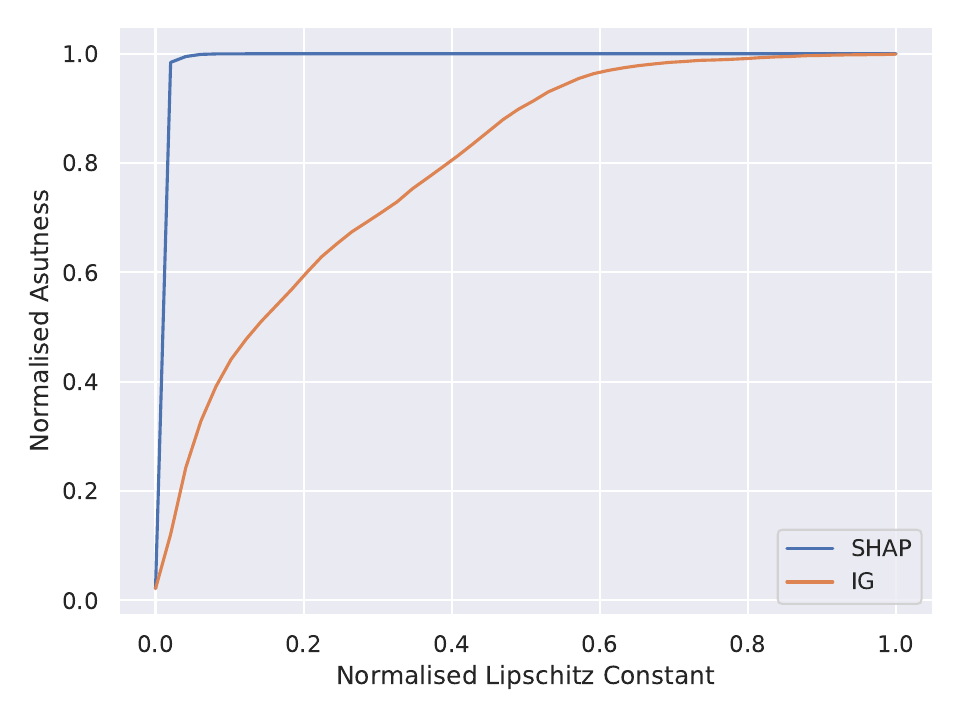}
  \caption{SHAP and Integrated Gradients normalised Astuteness curves on XOR problem.}
  \label{fig:xor_shap_ig_astute}
\end{figure}
\begin{figure}[t]
  \centering
  \includegraphics[scale = 0.5]{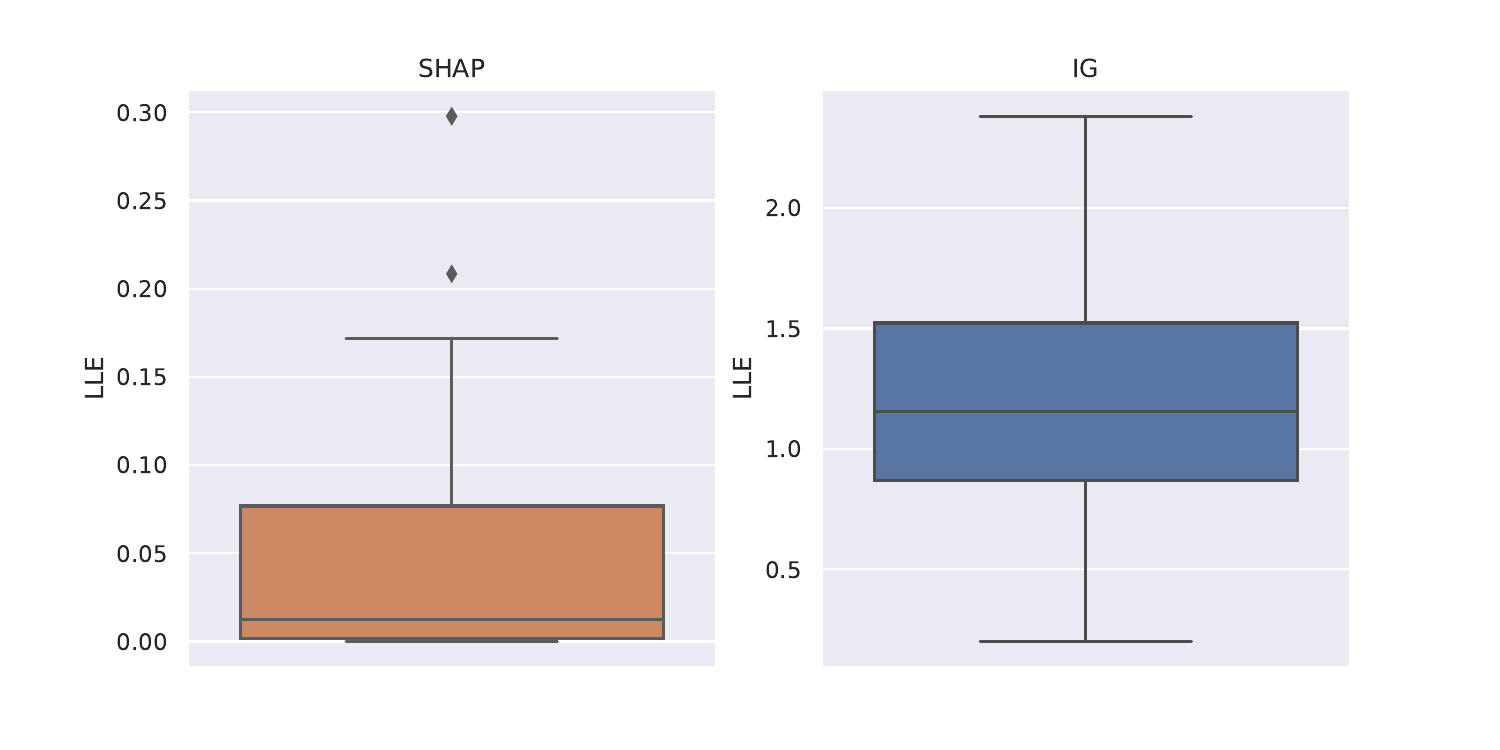}
  \caption{SHAP and Integrated Gradients Local Lipschitz Estimate distributions on XOR problem.}
  \label{fig:lle_box}
\end{figure}
\begin{table}[t]
\centering
\caption{Normalised astuteness AUC, LLE and AS of a two layer MLP on XOR ($r = \varepsilon = 0.26$).}
\begin{tabular}[t]{lll}
\toprule
Robustness Metric & Integrated Gradients & SHAP \\
\midrule
Normalised Astuteness AUC & $0.6969$ & $0.7819
$\\
Mean LLE  & $1.1952$ & $0.0414$\\
Mean AS &$0.9098$& $0.7564$ \\
\bottomrule
\label{table:table_1}
\end{tabular}
\end{table}%

\begin{table}[!ht]
\centering
\caption{Normalised Astuteness AUC of SHAP, LIME, and Integrated Gradients applied to XOR, Iris, and MNIST datasets.}
\label{table:table_2}
\begin{tabular}[]{llllll}
\toprule
Dataset & XAI Method & 2-Layer & 4-layer & 8-layer & 16-layer \\ \midrule
\multicolumn{1}{l|}{} & SHAP &$0.7819$ & $0.8491$ & $0.7483$ & $0.8356$ \\
\multicolumn{1}{l|}{XOR} & LIME & $\textbf{0.8899}$ & $\textbf{0.9033}$ & $\textbf{0.8747}$ & $\textbf{0.9078}$\\
\multicolumn{1}{l|}{} & IG &  $0.7614$ & $0.7967$ & $0.6266$ & $0.7659$ \\ \midrule
\multicolumn{1}{l|}{} & SHAP & $\textbf{0.9673}$ & $\textbf{0.9682}$ & $\textbf{0.9901}$ & $\textbf{0.9775}$\\
\multicolumn{1}{l|}{Iris} & LIME & $0.9365$ & $0.9025$ & $0.9865$ & $0.9710$ \\
\multicolumn{1}{l|}{} & IG &$0.7857$ & $0.9395$ & $0.8362$ & $0.8616$\\ \midrule
\multicolumn{1}{l|}{} & SHAP &$\textbf{0.9503}$ & $\textbf{0.9511}$ & $\textbf{0.9521}$ & $\textbf{0.9681}$\\
\multicolumn{1}{l|}{MNIST} & LIME &$0.6770$ & $0.6746$ & $0.7544$ & $0.7450$\\
\multicolumn{1}{l|}{} & IG &$0.7168$ & $0.6768$ & $0.6983$ & $0.7042$\\ \bottomrule
\end{tabular}
\end{table}%
In Figure \ref{fig:fig1}, explanations generated by SHAP are consistently similar for close by points. Normalised astuteness captures the robustness of SHAP and Integrated Gradients. In Table \ref{table:table_1}, SHAP and Integrated Gradients have a normalised astuteness AUC of 0.9895 and 0.7943 respectively. In Figure \ref{fig:xor_shap_ig_astute}, the normalised astuteness of SHAP converges rapidly for small Lipschitz constant. Hence for SHAP, the probability that close by points have similar explanations is higher than Integrated Gradients. In Figure \ref{fig:lle_box}, we have the distributions of LLE for SHAP and Integrated Gradients on XOR. LLE agrees with normalised astuteness that SHAP has a lower local Lipschitz constant than Integrated Gradients. We note from Figure \ref{fig:lle_box}, that Integrated Gradients has a symmetrical distribution of local Lipschitz constants with a mean LLE of $1.1952$. SHAP has a left-skewed distribution with a mean LLE of $0.0414$. The LLE of Integrated Gradients is consistent over points. The outliers in SHAP's LLE may be due to points close to the decision boundary where opposite classes may be in the local neighbourhood of a point. 

Whilst LLE, AS and normalised astuteness agree that SHAP is more robust than Integrated Gradients, the usability and interpretation of the metrics is where they differ. Normalised astuteness provides a single number to quantify the probability of local points having similar explanations. Whilst LLE and AS provide a similar measure, both are not quantified into a single metric. Furthermore, astuteness provides an assurance in probability of the robustness. Whereas, LLE and AS do not. Furthermore, LLE and AS do not provide a measure of perfect robustness. Whilst a normalised astuteness of 1 indicates perfect robustness.

We evaluate normalised astuteness on XOR, Iris, and MNIST, three well known machine learning datasets. We train MLPs of  2, 4, 8, and 16 layers with ReLU activation. We take $r \geq 0$ as the median point-wise distance between points in the dataset as in \cite{khan2023analyzing}. 

In Table \ref{table:table_2}, SHAP or LIME has a higher robustness than Integrated Gradients. The choice of base-point significantly impacts the similarity of explanations generated for close by points \cite{wang2020smoothed}. We leave a further analysis of the impact of the base-point choice and Integrated Gradients robustness to future work.

\section{The stable rank as a measure of robustness}
Here we extend the link between stable rank and the Lipschitz constant to explainability models. Specifically we demonstrate a theoretical lower bound of the Lipschitz constant which depends on the stable rank of the embedding matrix. We then extend the use of the stable rank to probabilistic Lipschitzness and show the efficacy of the stable rank as a heuristic for explainer astuteness. 

Consider a $k$-layer MLP as the following composition of functions
        \begin{equation*}
            f(x) = (g_k \circ \psi_{k-1} \circ g_{k} \circ \ldots \circ \psi_{1} \circ g_{1})(x),
        \end{equation*}
        where, $g_{i}(x) = \mathbf{W}_{i}x+b_{i}$ is an affine projection with trained weight matrix $\mathbf{W}_{i} \in \mathbb{R}^{n \times m}$, $b_{i} \in \mathbb{R}^{n}$ is the bias, and $\psi_{i}$ is a non-linear activation function. 
In \cite{NIPS2017_b22b257a, pmlr-v40-Neyshabur15} the authors prove the following generalisation bound for $k$-layer neural networks,
\begin{equation*}
\mathcal{O}\sqrt{\prod_{i}^{k}\|\mathbf{W}_{i}\|^{2}_{2}\sum_{i}^{k}S(\mathbf{W}_{i})},
\end{equation*}
where
\begin{equation*}
 S(\mathbf{W}) = \frac{\|\mathbf{W} \|_{F}}{\| \mathbf{W} \|_{2}}  
\end{equation*}
is the stable rank, a lower bound approximation of the rank of a matrix. For a neural network $f$, the Lipschitz constant is bounded above by
 \begin{equation*}
     L \leq \prod_{i = 1}^N L(\phi_{i}),
 \end{equation*}
 where $\phi_{i}$ is an affine projection $g_{i}$ or a non-linear activation function $\psi_{i}$. For the $\ell_2$ norm, $L(g_{i})$ is the spectral norm of the weight matrix. Common activation functions are 1-Lipschitz. Hence the Lipschitz constant upper bound becomes.
 \begin{equation*}
     L \leq \prod_{i = 1}^N \|\mathbf{W}_{i}\|_{2}.
 \end{equation*}
 The generalisation bound of \cite{NIPS2017_b22b257a, pmlr-v40-Neyshabur15} is therefore dependent on the stable rank of the weight matrices and the Lipschitz constant. The interplay of the stable rank and Lipschitz constant is investigated in \cite{ramasinghe2022periodicity, sanyal2020stable}. In \cite{sanyal2020stable}, optimising both the stable rank and Lipschitz constant is shown to improve generalisation and classification performance. In \cite{ramasinghe2022periodicity} the authors investigate the relation between the stable rank and Lipschitz constant in signal reconstruction. 

Given a neural network $f$, let $\phi$ be the \textit{k}-th layer of the neural network. Suppose we have $N$ training points $\{x_{i}\}_{i=1}^{N}$. Evaluating $\phi$ on each training point we have the following embedding matrix.
        \begin{equation*}
            \mathbf{X} \in \mathbb{R}^{D \times N} \coloneqq [\phi(x_{1})^{T} \: \phi(x_{2})^{T} \ldots \: \phi(x_{N})^{T}].
        \end{equation*}
        
Demonstrated empirically in \cite{ramasinghe2022periodicity}, the magnitude of the stable rank of the embedding matrix $\mathbf{X}$ is related to the magnitude of the Lipschitz constant. As the Lipschitz constant is hard to calculate \cite{scaman2019lipschitz, jordan2021exactly} the stable rank of the embedding matrix can act as a heuristic of the Lipschitz constant. The relationship between the stable rank and Lipschitz constant in \cite{ramasinghe2022periodicity} is investigated for two special cases of the embedding matrix, $\mathbf{X}$, not in full generality.
        
\subsection{Stable rank and the Lipschitz constant}
In this section, a theoretical lower bound of the Lipschitz constant dependent on the stable rank of the embedding matrix is established. This bound works towards a theoretical justification of the stable rank, Lipschitz relationship claimed in \cite{ramasinghe2022periodicity}.

Suppose $\phi$ is $L$-Lipschitz. 
      Given $n$ training points $\{x_{i}\}_{i}^{n}$ we define the matrices $\mathbf{D},\mathbf{Y} \in \mathbb{R}^{n\times n}$ and $\mathbf{X} \in \mathbb{R}^{n \times m}$ as
      \begin{align*}
          &\mathbf{D}_{ij}  = \|\phi(x_i)-\phi(x_j)\|^{2}\\
          &\mathbf{Y}_{ij}  =\|x_{i}-x_{j} \|^{2}\\
          &\mathbf{X}  = [\phi(x_{1}), \: \phi(x_{2}),
            \ldots , \phi(x_{n})]^{T}.\\
      \end{align*}
      Clearly,
      \begin{equation*}
          \mathbf{D}_{ij} \leq L^{2} \mathbf{Y}_{ij}.
      \end{equation*}
      We can write $\mathbf{D}$ in the following matrix form
      \begin{equation*}
          \mathbf{D} = \mathbf{1}\cdot\diag(\mathbf{X}^T\mathbf{X})+\diag(\mathbf{X}^T\mathbf{X})\cdot \mathbf{1} -2\mathbf{X}^T\mathbf{X}.
      \end{equation*}
      Now let 
      \begin{equation*}
      \mathbf{M} = \mathbf{1}\cdot\diag(\mathbf{X}^T\mathbf{X})+
      \diag(\mathbf{X}^T\mathbf{X})\cdot \mathbf{1}.
      \end{equation*}
      Consider the matrices 
      \begin{equation*}
        \mathbf{E}(n)_{ij} = \left \{
        \begin{aligned}
            &2, && \text{if}\ i = n = j\\
            &1, && \text{if}\ i = n \ \text{or}\ j=n \\
            &0, && \text{otherwise},
          \end{aligned}\right. 
      \end{equation*} 
      then,
      \begin{equation*}
          \mathbf{M} = \sum_{i} (\mathbf{X}^T\mathbf{X})_{ii}\mathbf{E}(i).
      \end{equation*}

      Now looking at the Frobenius norm of $\mathbf{Y}$ and $\mathbf{D}$. It follows from the fact that both $\mathbf{D}$ and $\mathbf{Y}$ are symmetric and
        \begin{equation*}
          \mathbf{D}_{ij} \leq L^{2} \mathbf{Y}_{ij}
      \end{equation*}
      that,
      \begin{equation*}
          \|\mathbf{D} \|_{F}^{2} \leq L^4 \| \mathbf{Y} \|_{F}^{2}.
      \end{equation*}

      As $\mathbf{D}$ is the sum of matrices we have an explicit formula for $\|\mathbf{D}\|_{F}^{2}$,
      \begin{equation*}
          \|\mathbf{D}\|_{F}^{2} = \| \mathbf{M} \|_{F}^{2} + 2\|\mathbf{X}^{T}\mathbf{X} \|_{F}^{2} -4\langle \mathbf{M}, \mathbf{X}^{T}\mathbf{X} \rangle_{F}.
      \end{equation*}
      Computing $\|\mathbf{M}\|_{F}^{2}$ we have,
      
      \begin{equation*}
          \|\mathbf{M}\|_{F}^{2} = \|\mathbf{1}\cdot \diag(\mathbf{X}^{T}\mathbf{X})\|_{F}^{2} +\|\diag(\mathbf{X}^{T}\mathbf{X})\cdot\mathbf{1} \|_{F}^{2}+ 2\langle \mathbf{1}\cdot \diag(\mathbf{X}^{T}\mathbf{X}), \diag(\mathbf{X}^{T}\mathbf{X})\cdot\mathbf{1} \rangle_{F}.
      \end{equation*}
      
      Calculating the Frobenius inner product
      \begin{align*}
          \langle \mathbf{1}\cdot \diag(\mathbf{X}^{T}\mathbf{X}), \diag(\mathbf{X}^{T}\mathbf{X})\cdot\mathbf{1} \rangle_{F} &= \tr((\mathbf{1}\cdot \diag(\mathbf{X}^{T}\mathbf{X}))^{T} \diag(\mathbf{X}^{T}\mathbf{X})\cdot\mathbf{1})\\
          &=\tr((\diag(\mathbf{X}^{T}\mathbf{X})\cdot\mathbf{1})^2)\\
          &=\tr(\tr(\diag(\mathbf{X}^{T}\mathbf{X}))\mathbf{1}\cdot \diag(\mathbf{X}^{T}\mathbf{X}))\\
          &=\tr(\diag(\mathbf{X}^{T}\mathbf{X}))\tr(\mathbf{1}\cdot \diag(\mathbf{X}^{T}\mathbf{X}))\\
          &=\tr(\diag(\mathbf{X}^{T}\mathbf{X}))\tr( \diag(\mathbf{X}^{T}\mathbf{X}))\\
          &=\tr(\mathbf{X}^{T}\mathbf{X})^{2}\\
          &= \|\mathbf{X}\|_{F}^{4}.
      \end{align*}
      It is easy to show that $\|\mathbf{1}\cdot \diag(\mathbf{X}^{T}\mathbf{X})\|_{F}^{2} = \| \diag(\mathbf{X}^{T}\mathbf{X})\cdot\mathbf{1}\|_{F}^{2}$. Now calculating $\|\mathbf{1}\cdot \diag(\mathbf{X}^{T}\mathbf{X})\|_{F}^{2}$.
      
      \begin{align*}
         \|\mathbf{1}\cdot \diag(\mathbf{X}^{T}\mathbf{X})\|_{F}^{2} &= \tr( (\mathbf{1}\cdot \diag(\mathbf{X}^{T}\mathbf{X}))^{T}(\mathbf{1}\cdot \diag(\mathbf{X}^{T}\mathbf{X})))\\
         &= \tr(\diag(\mathbf{X}^{T}\mathbf{X})(\mathbf{1}\cdot \mathbf{1})\diag(\mathbf{X}^{T}\mathbf{X}))\\
         &=n\tr(\diag(\mathbf{X}^{T}\mathbf{X})^2)\\
         &= n\sum_{i}\mathbf{X}^{T}\mathbf{X}_{ii}^2.
      \end{align*}
      Hence,
      \begin{equation*}
          \|\mathbf{M}\|_{F}^{2} = 2n\sum_{i} \mathbf{X}^{T}\mathbf{X}_{ii}^{2}+2\|\mathbf{X}\|_{F}^{4}.
      \end{equation*}
      Noting that
      \begin{equation*}
          \mathbf{M} = \sum_{i}(\mathbf{X}^{T}\mathbf{X})_{ii}\mathbf{E}(i),
      \end{equation*}
      we have
      \begin{align*}
          \langle \mathbf{M}, \mathbf{X}^{T}\mathbf{X} \rangle_{F} &= \sum_{i}(\mathbf{X}^{T}\mathbf{X})_{ii} \langle \mathbf{E}(i), \mathbf{X}^{T}\mathbf{X} \rangle_{F}\\
          &=2\sum_{i}(\mathbf{X}^{T}\mathbf{X})_{ii}\sum_{j}(\mathbf{X}^{T}\mathbf{X})_{ji}.
      \end{align*}
      
      So we have
      \begin{align*}
          \|\mathbf{D}\|_{F}^{2} &= 4n\sum_{i} \mathbf{X}^{T}\mathbf{X}_{ii}^{2}+2\|\mathbf{X}\|_{F}^{4} - 4\langle \mathbf{M}, \mathbf{X}^{T}\mathbf{X} \rangle_{F}\\
          &=4(n-2)\sum_{i} (\mathbf{X}^{T}\mathbf{X})_{ii}^{2}-8\sum_{i}(\mathbf{X}^{T}\mathbf{X})_{ii}\sum_{j \neq i}(\mathbf{X}^{T}\mathbf{X})_{ji}+2\|\mathbf{X}\|_{F}^{4}\\
          &=4(n-2)\sum_{i} (\mathbf{X}^{T}\mathbf{X})_{ii}^{2}-8\sum_{i}(\mathbf{X}^{T}\mathbf{X})_{ii}\sum_{j \neq i}(\mathbf{X}^{T}\mathbf{X})_{ji}+2\|\mathbf{X}\|_{2}^{4}S(\mathbf{X})^4.
      \end{align*}
      We have that
      \begin{align*}
          \sum_{i}(\mathbf{X}^{T}\mathbf{X})_{ii}\sum_{j \neq i}(\mathbf{X}^{T}\mathbf{X})_{ji} = \tr(\mathbf{J}\mathbf{X}^{T}\mathbf{X}\diag(\mathbf{X}^{T}\mathbf{X}))
      \end{align*}
      where $\mathbf{J}$ is the matrix with 0's on the diagonal and 1's elsewhere. It follows that,
      
      \begin{equation*}
          \|\mathbf{D}\|^{2}_{F} = 4(n-2)\|\diag(\mathbf{X}^{T}\mathbf{X})\|^{2}_{2}S(\diag(\mathbf{X}^{T}\mathbf{X}))^{2}-8\tr(\mathbf{J}\mathbf{X}^{T}\mathbf{X}\diag(\mathbf{X}^{T}\mathbf{X})) + 2\|\mathbf{X}\|_{2}^{4}S(\mathbf{X})^4.
      \end{equation*}

      We note from above that the Lipschitz constant of a classifier is bounded above by the product of spectral norms of the weight matrices.
      
      \begin{equation*}
       \frac{\|\mathbf{D} \|_{F}^{2}}{\| \mathbf{Y} \|_{F}^{2}} \leq L^4 \leq \prod_{i = 1}^N \|\mathbf{W}_{i}\|_{2}^{4}.
      \end{equation*}

      The inequality above is a step towards proving a relationship between the stable rank and Lipschitz constant for a general embedding matrix. In Sections 4 and 5 we demonstrate the claim in \cite{khan2023analyzing} that smooth classifiers result in higher astute explainers. We now make the following claim: Large stable rank results in low astuteness and consequentially less robust explainers.
      
    \subsection{Stable Rank and Astuteness}
      In this section, we demonstrate the relationship between astuteness and stable rank. First, the relationship between stable rank and neural network astuteness is demonstrated on MNIST image reconstruction with autoencoders. We demonstrate our theoretical lower bound for the Lipschitz constant holds on this problem as we can exactly compute the Lipschitz constant. Second, we demonstrate the stable rank as a heuristic for explainer astuteness.

\subsubsection{Stable Rank and Neural Network Astuteness}
      
\begin{figure}[t]
  \centering
  \includegraphics[scale = 0.7]{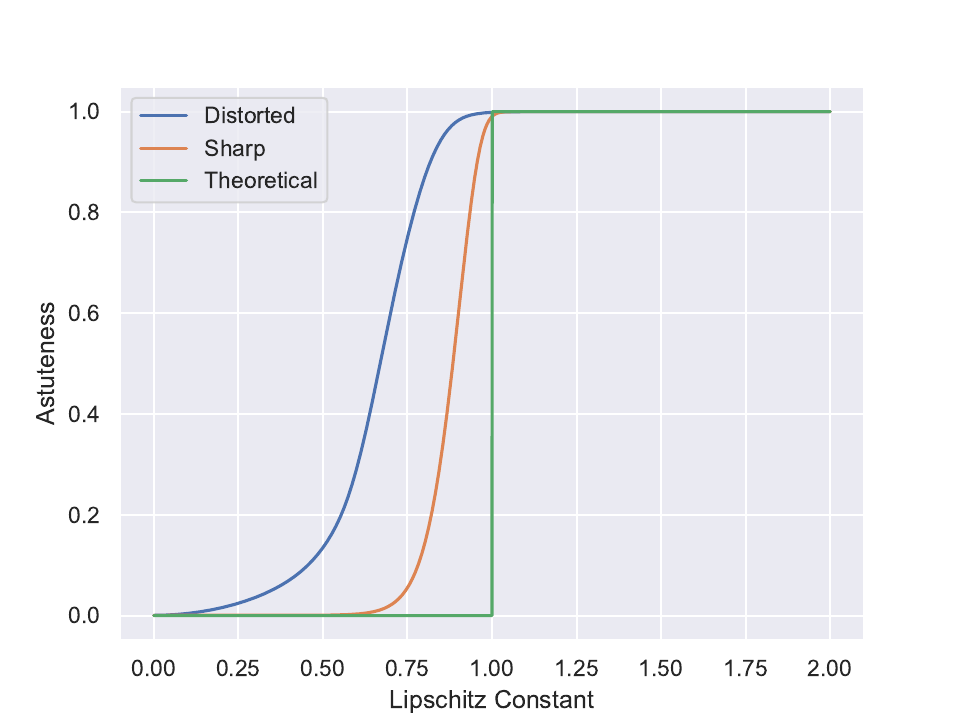}
  \caption{Astuteness curves for autoencoders with Gaussian activations trained on MNIST.}
  \label{fig:fig5}
\end{figure}

\begin{figure}[t]
  \centering
  \includegraphics[scale = 0.7]{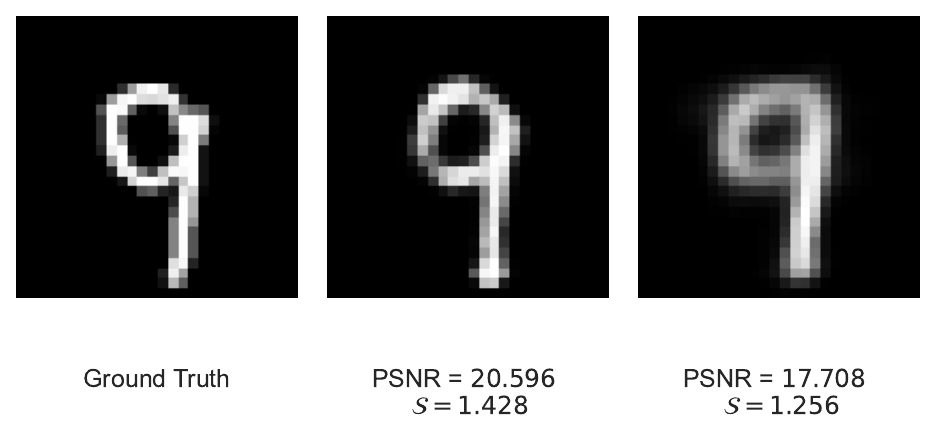}
  \caption{Autoencoder predictions with Peak Signal-to-Noise Ratio (PSNR) and stable rank $\mathcal{S}$. \textbf{Left}: Ground Truth, \textbf{Middle}: Sharp Autoencoder, \textbf{Right}: Distorted Autoencoder.}
  \label{fig:fig2}
\end{figure}
      Suppose we have an autoencoder $f$, trained on MNIST data. Let the embedding matrix $\mathbf{X}$ be the reconstructions on the input image $x$. Given a perfectly trained autoencoder $f(x) = x$,  the Lipschitz constant of the output layer is 1. We expect the astuteness curve of $f$ to converge to 1 at $L = 1$. To verify this experimentally we train two autoencoders on MNIST with Gaussian activation. For $a \in \mathbb{R}\setminus \{0\}$, Gaussian activation \cite{ramasinghe2022periodicity} is defined as
      \begin{equation*}
          g(x) = \exp(\frac{-0.5x^2}{a^2}).
      \end{equation*}
      We alter the parameter of the Gaussian activation to achieve distorted and sharp reconstructions. In Figure \ref{fig:fig5} the autoencoders astuteness curves converge at 1. The distorted autoencoder in Figure \ref{fig:fig5} converges faster, as distorted reconstructions have lower Lipschitz constant \cite{ramasinghe2022periodicity}. For image reconstructions, the peak signal-to-noise- ratio (PSNR) corresponds to reconstruction smoothness and therefore the Lipschitz constant of the autoencoder. Sharp reconstructions have a higher PSNR, corresponding to a higher lipschitz constant and stable rank \cite{ramasinghe2022periodicity}. In Figure \ref{fig:fig2} we demonstrate a distorted and sharp reconstruction with low and high PSNRs and have low and high stable ranks respectively. For a perfectly trained autoencoder the Lipschitz constant is 1. Hence from the inequality in Section 6.1 we have Lipschitz lower bound $L_{LB} \leq 1$. For the sharp and distorted autoencoder we have Lipschitz lower bounds $L^{\text{SHARP}}_{\text{LB}} = 0.9109$, $L^{\text{DIST}}_{\text{LB}} = 0.7548$ and stable ranks $\mathcal{S}^{\text{SHARP}} = 1.428$, $\mathcal{S}^{\text{DIST}} = 1.256$,  respectively. The lower bounds match the observations in Figure \ref{fig:fig5}, for a sharp reconstruction the astuteness is lower due to a higher Lipschitz constant.

\subsubsection{Stable Rank and Explanation Astuteness}
\begin{figure}[t]
  \centering
  \includegraphics[scale = 0.6]{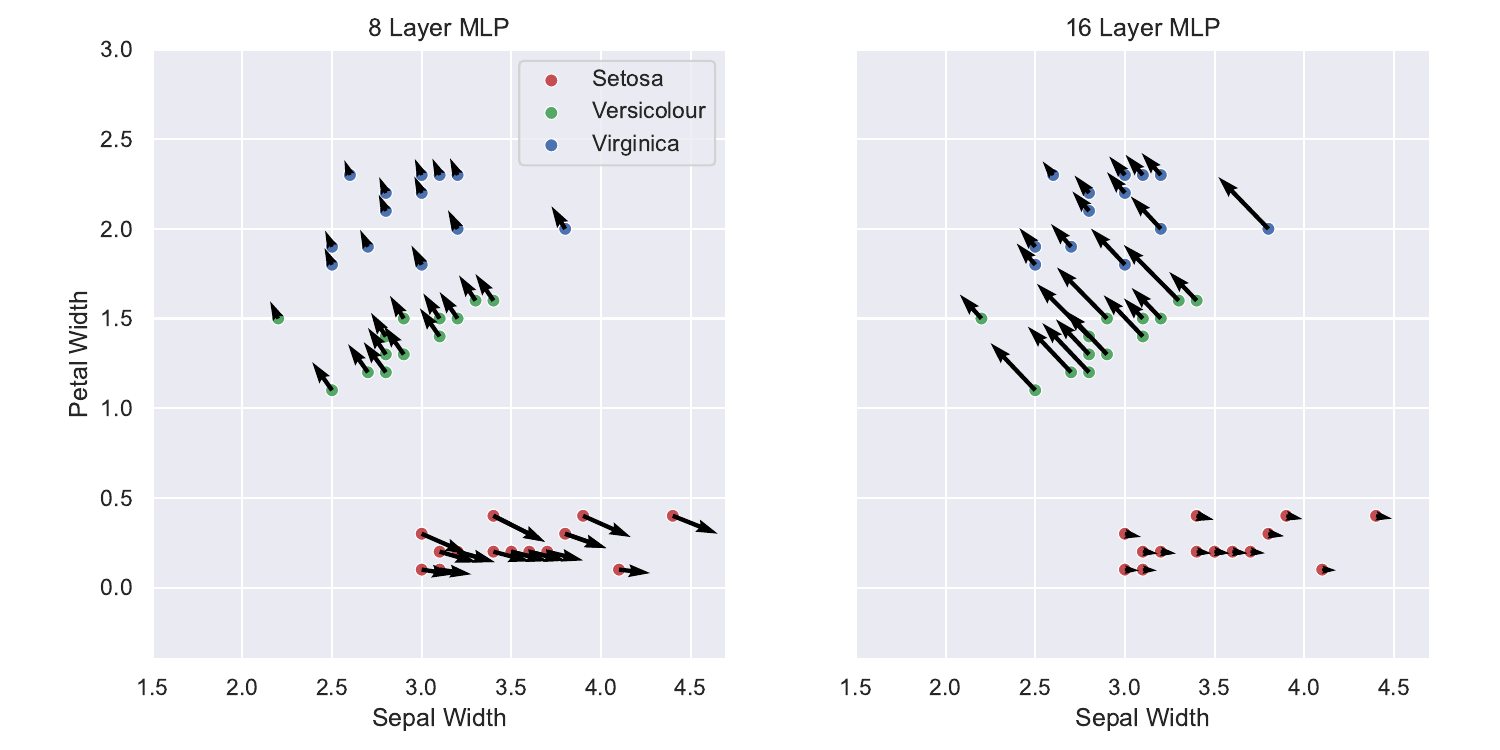}
  \caption{Integrated Gradients explanations for 8-and 16-Layer MLP on Iris.}
  \label{fig:sepal_exps}
\end{figure}

\begin{table}[t]
\centering
\caption{Normalised astuteness AUC and stable rank of 8-and 16-layer MLP on Iris.}
\label{table:table_3}
\begin{tabular}[t]{lll}
\toprule
 & 8-Layer & 16-Layer \\
\midrule
Stable Rank  & $1.2935$ & $1.2969$\\
Classifier Astuteness AUC & $0.9189
$ & $0.8146$\\
Integrated Gradients AUC & $0.9733
$ & $0.8921$\\
\bottomrule
\end{tabular}
\end{table}%

\begin{table}[t]
\centering
\caption{Normalised MLP Astuteness AUC and stable rank of 2 and 4 layer MLP on MNIST.}
\label{table:table_4}
\begin{tabular}[t]{lll}
\toprule
 & 2-Layer & 4-Layer \\
\midrule
Stable Rank  & $2.9098$ & $2.5214$\\
Classifier Astuteness AUC & $0.8828
$ & $0.9127$\\
\bottomrule
\end{tabular}
\end{table}%

\begin{figure}[!ht]
  \centering
  \includegraphics[scale = 0.6]{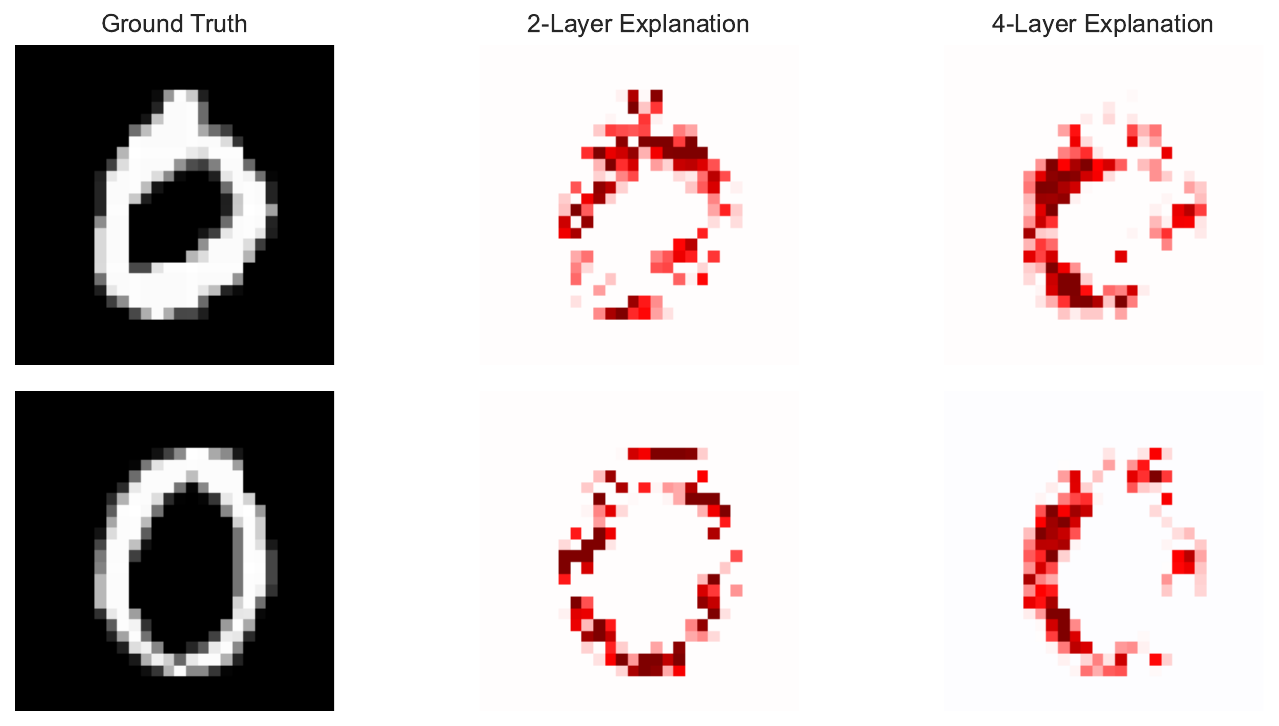}
  \caption{Integrated Gradients explanations of two MNIST zero digits from a 2-and 4-Layer MLP.}
  \label{fig:ig_mnist}
\end{figure}

Results in Section 4 and \cite{khan2023analyzing} have established the link between the Lipschitz constant of the classifier and the robustness of explainability models. The link both theoretically and experimentally between the stable rank and the Lipschitz constant of classifiers has been established.  Classifier astuteness and stable rank are therefore related. We now make the following claim akin to \cite{ramasinghe2022periodicity}. The stable rank can be used as a heuristic for the robustness of explainability models. 

The efficacy of stable rank as a robustness heuristic is demonstrated on three classification problems. Firstly, an 8-and 16-layer MLP with ReLU activation is trained on the Iris dataset. Secondly a two 2-layer MLPs with ReLU and tanh activation is trained on the XOR problem. Lastly, a 2-and 4-layer MLP is trained on MNIST. The different layers and activation functions are used to demonstrate the efficacy of stable rank as a heuristic under differing neural architectures.

\subsubsection{Stable rank under differing activation functions}

Two MLPs with $\ReLU$ and tanh activations respectively are trained on the XOR problem. In Table \ref{table:table_3} we have the stable rank and normalised astuteness AUC. The stable rank of the ReLU MLP is larger than the tanh MLP and therefore the Lipschitz constant of the ReLU is Larger than the tanh MLP. The astuteness AUC of $\ReLU$ is lower than tanh, corresponding to a larger Lipschitz constant and stable rank. Explanations generated for the $\ReLU$ MLP are less robust than the tanh explanations. Thus, for the tanh MLP close by points have similar explanations.

\begin{table}[t]
\centering
\caption{Normalised astuteness AUC and stable rank of 2-layer MLP on XOR problem.}
\label{table:table_5}
\begin{tabular}[t]{lll}
\toprule
 & $\ReLU$ & tanh \\
\midrule
Stable Rank  & $1.6805$ & $1.1527$\\
Astuteness AUC & $0.6304$ & $0.7403$\\
Integrated Gradients AUC &$0.7614$& $0.8616$ \\
SHAP AUC &$0.6248$ & $0.7441$\\
\bottomrule
\end{tabular}
\end{table}%

\subsubsection{Stable rank under differing layers}

For another example, an 8-and 16-layer MLP with ReLU activation is trained on two features of the Iris dataset. In Table~\ref{table:table_4} the normalised classifier astuteness AUC of the 8-layer MLP is higher than the 16-layer MLP. Consequentially the stable rank of the 8-layer is lower than the 16-layer. In Figure \ref{fig:sepal_exps}, for the 8-Layer MLP close by points consistently have similar explanation vectors (in both magnitude and direction). For the 16-layer MLP the Setosa class has explanations that are approximately the same in comparison to the 8-layer Setosa explanations. The similarity of the 16-layer Setosa explanation vectors is countered by the Versicolour and Virginica explanations vectors. Nearby points of both Versicolour and Virginica have explanations that differ significantly in magnitude. In Table~\ref{table:table_4} the stable rank of the 8-layer MLP is lower then the 16-layer. Consequently, the 8-layer MLP astuteness AUC is higher than the 16-layer. The normalised AUC of Integrated Gradients for the 8-layer MLP is higher than the 16-layer as a smoother classifier provides more robust explanations.

For a final example, a 2-and 4-layer MLP is trained on MNIST. In Table~\ref{table:table_5} the 4-Layer MLP has a higher stable rank and therefore lower astuteness. The explanations of the 4-layer should therefore have a higher robustness than the 16-layer. In Figure \ref{fig:ig_mnist} Integrated Gradients for the 4-layer MLP provides explanations that are similar for close by points, in this case two images of handwritten zeros.

The magnitude of the stable rank therefore provides insight into the robustness of explainability techniques. Furthermore, the stable rank can be applied directly to explainability models. We leave the analysis of the stable rank to explainability models for future work.  

\section{Conclusions and Future Work}
In this paper, we proved theoretical lower bound guarantees on the probabilistic Lipschitzness of Integrated Gradients, LIME, and SmoothGrad. The theoretical lower bounds further the claim in \cite{khan2023analyzing} that the smoothness of a neural network determines the robustness of the explainability models. We introduced normalised astuteness as a metric to compare explainability techniques. Normalised astuteness demonstrated across three common machine learning datasets that integrated gradients provides explanations of a lower robustness than SHAP or LIME. Additionally, we demonstrated that the stable rank can be used as a heuristic measure of explainability robustness and proved a relationship between the stable rank and Lipschitz constant of a classifier. The stable rank was demonstrated to agree with normalised astuteness, however, further investigation is required to establish the efficacy of the stable rank as a robustness metric. In future work we seek to apply normalised astuteness to graph neural networks and extend the theoretical astuteness and stable rank results to graph explainability models. 

\section*{Acknowledgements}
The Commonwealth of Australia (represented by the Defence Science and Technology Group) supports this research through a Defence Science Partnerships agreement. Lachlan Simpson is supported by a University of Adelaide scholarship.

\bibliographystyle{IEEEtran}

\bibliography{references}

\begin{thebibliography}{10}
\providecommand{\url}[1]{#1}
\csname url@samestyle\endcsname
\providecommand{\newblock}{\relax}
\providecommand{\bibinfo}[2]{#2}
\providecommand{\BIBentrySTDinterwordspacing}{\spaceskip=0pt\relax}
\providecommand{\BIBentryALTinterwordstretchfactor}{4}
\providecommand{\BIBentryALTinterwordspacing}{\spaceskip=\fontdimen2\font plus
\BIBentryALTinterwordstretchfactor\fontdimen3\font minus
  \fontdimen4\font\relax}
\providecommand{\BIBforeignlanguage}[2]{{%
\expandafter\ifx\csname l@#1\endcsname\relax
\typeout{** WARNING: IEEEtran.bst: No hyphenation pattern has been}%
\typeout{** loaded for the language `#1'. Using the pattern for}%
\typeout{** the default language instead.}%
\else
\language=\csname l@#1\endcsname
\fi
#2}}
\providecommand{\BIBdecl}{\relax}
\BIBdecl

\bibitem{yolo}
J.~Redmon, S.~Divvala, R.~Girshick, and A.~Farhadi, ``You only look once:
  Unified, real-time object detection,'' \emph{Proceedings of the IEEE
  Conference on Computer Vision and Pattern Recognition (CVPR)}, 2016.

\bibitem{Liu_2023}
Y.~Liu, T.~Han, S.~Ma, J.~Zhang, Y.~Yang, J.~Tian, H.~He, A.~Li, M.~He, Z.~Liu,
  Z.~Wu, L.~Zhao, D.~Zhu, X.~Li, N.~Qiang, D.~Shen, T.~Liu, and B.~Ge,
  ``Summary of {ChatGPT}-related research and perspective towards the future of
  large language models,'' \emph{Meta-Radiology}, vol.~1, no.~2, p. 100017,
  2023.

\bibitem{9737249}
K.~Millar, L.~Simpson, A.~Cheng, H.~G. Chew, and C.-C. Lim, ``Detecting botnet
  victims through graph-based machine learning,'' \emph{2021 International
  Conference on Machine Learning and Cybernetics (ICMLC)}, pp. 1--6, 2021.

\bibitem{alphafold}
J.~Jumper, R.~Evans, A.~Pritzel, T.~Green, M.~Figurnov, O.~Ronneberger,
  K.~Tunyasuvunakool, R.~Bates, A.~{\v Z}{\'\i}dek, A.~Potapenko, A.~Bridgland,
  C.~Meyer, S.~A.~A. Kohl, A.~J. Ballard, A.~Cowie, B.~Romera-Paredes,
  S.~Nikolov, R.~Jain, J.~Adler, T.~Back, S.~Petersen, D.~Reiman, E.~Clancy,
  M.~Zielinski, M.~Steinegger, M.~Pacholska, T.~Berghammer, S.~Bodenstein,
  D.~Silver, O.~Vinyals, A.~W. Senior, K.~Kavukcuoglu, P.~Kohli, and
  D.~Hassabis, ``Highly accurate protein structure prediction with alphafold,''
  \emph{Nature}, vol. 596, no. 7873, pp. 583--589, 2021.

\bibitem{Sejnowski_2020}
T.~J. Sejnowski, ``The unreasonable effectiveness of deep learning in
  artificial intelligence,'' \emph{Proceedings of the National Academy of
  Sciences}, vol. 117, no.~48, pp. 30\,033--30\,038, 2020.

\bibitem{zednik2019solving}
C.~Zednik, ``\BIBforeignlanguage{English}{Solving the black box problem: A
  normative framework for explainable artificial intelligence},''
  \emph{\BIBforeignlanguage{English}{Philosophy \& Technology}}, vol.~34,
  no.~2, pp. 265--288, 2021.

\bibitem{dong2019efficient}
Y.~Dong, H.~Su, B.~Wu, Z.~Li, W.~Liu, T.~Zhang, and J.~Zhu, ``Efficient
  decision-based black-box adversarial attacks on face recognition,''
  \emph{IEEE/CVF Conference on Computer Vision and Pattern Recognition (CVPR)},
  pp. 7706--7714, 2019.

\bibitem{8578273}
K.~Eykholt, I.~Evtimov, E.~Fernandes, B.~Li, A.~Rahmati, C.~Xiao, A.~Prakash,
  T.~Kohno, and D.~Song, ``Robust physical-world attacks on deep learning
  visual classification,'' \emph{IEEE/CVF Conference on Computer Vision and
  Pattern Recognition (CVPR)}, pp. 1625--1634, 2018.

\bibitem{szegedy2014intriguing}
C.~Szegedy, W.~Zaremba, I.~Sutskever, J.~Bruna, D.~Erhan, I.~Goodfellow, and
  R.~Fergus, ``Intriguing properties of neural networks,'' \emph{arXiv preprint
  arXiv:1312.6199}, 2013.

\bibitem{ieeeXAIsurvey}
E.~Tjoa and C.~Guan, ``A survey on explainable artificial intelligence (xai):
  Toward medical xai,'' \emph{IEEE Transactions on Neural Networks and Learning
  Systems}, vol.~32, no.~11, pp. 4793--4813, 2021.

\bibitem{agarwal2021unification}
S.~Agarwal, S.~Jabbari, C.~Agarwal, S.~Upadhyay, S.~Wu, and H.~Lakkaraju,
  ``Towards the unification and robustness of perturbation and gradient based
  explanations,'' \emph{Proceedings of the 38th International Conference on
  Machine Learning}, vol. 139, pp. 110--119, 2021.

\bibitem{ieeeGraphXAI}
D.~Bacciu and D.~Numeroso, ``Explaining deep graph networks via input
  perturbation,'' \emph{IEEE Transactions on Neural Networks and Learning
  Systems}, vol.~34, no.~12, pp. 10\,334--10\,345, 2023.

\bibitem{ribeiro2016why}
M.~T. Ribeiro, S.~Singh, and C.~Guestrin, ``"why should i trust you?":
  Explaining the predictions of any classifier,'' \emph{Proceedings of the 22nd
  ACM SIGKDD International Conference on Knowledge Discovery and Data Mining},
  p. 1135–1144, 2016.

\bibitem{lundberg2017unified}
S.~M. Lundberg and S.-I. Lee, ``A unified approach to interpreting model
  predictions,'' \emph{Proceedings of the 31st International Conference on
  Neural Information Processing Systems (NeurIPS)}, pp. 4768--4777, 2017.

\bibitem{pmlr-v70-sundararajan17a}
M.~Sundararajan, A.~Taly, and Q.~Yan, ``Axiomatic attribution for deep
  networks,'' \emph{Proceedings of the 34th International Conference on Machine
  Learning (ICML)}, vol.~70, pp. 3319--3328, 2017.

\bibitem{smilkov2017smoothgrad}
D.~Smilkov, N.~Thorat, B.~Kim, F.~Viégas, and M.~Wattenberg, ``Smoothgrad:
  removing noise by adding noise,'' \emph{arXiv preprint arXiv:1706.03825},
  2017.

\bibitem{LAMY201942}
J.-B. Lamy, B.~Sekar, G.~Guezennec, J.~Bouaud, and B.~S{\'e}roussi,
  ``Explainable artificial intelligence for breast cancer: A visual case-based
  reasoning approach,'' \emph{Artificial Intelligence in Medicine}, vol.~94,
  pp. 42--53, 2019.

\bibitem{Zhang_2022}
Z.~Zhang, H.~A. Hamadi, E.~Damiani, C.~Y. Yeun, and F.~Taher, ``Explainable
  artificial intelligence applications in cyber security: State-of-the-art in
  research,'' \emph{{IEEE} Access}, vol.~10, pp. 93\,104--93\,139, 2022.

\bibitem{khan2023analyzing}
Z.~Khan, D.~Hill, A.~Masoomi, J.~Bone, and J.~Dy, ``Analyzing explainer
  robustness via lipschitzness of prediction functions,'' \emph{arXiv preprint
  arXiv:2206.12481}, 2023.

\bibitem{yeh2019infidelity}
C.-K. Yeh, C.-Y. Hsieh, A.~S. Suggala, D.~I. Inouye, and P.~Ravikumar, ``On the
  (in)fidelity and sensitivity of explanations,'' \emph{Proceedings of the 33rd
  International Conference on Neural Information Processing Systems}, 2019.

\bibitem{Montavon_2018}
G.~Montavon, W.~Samek, and K.-R. Müller, ``Methods for interpreting and
  understanding deep neural networks,'' \emph{Digital Signal Processing},
  vol.~73, pp. 1--15, 2018.

\bibitem{scaman2019lipschitz}
K.~Scaman and A.~Virmaux, ``Lipschitz regularity of deep neural networks:
  analysis and efficient estimation,'' \emph{Advances in Neural Information
  Processing Systems (NeurIPS)}, vol.~31, 2018.

\bibitem{jordan2021exactly}
M.~Jordan and A.~G. Dimakis, ``Exactly computing the local lipschitz constant
  of relu networks,'' \emph{Advances in Neural Information Processing Systems
  (NeurIPS)}, vol.~33, pp. 7344--7353, 2020.

\bibitem{ramasinghe2022periodicity}
S.~Ramasinghe and S.~Lucey, ``Beyond periodicity: Towards a unifying framework
  for activations in coordinate-mlps,'' \emph{European Conference on Computer
  Vision (ECCV)}, pp. 142--158, 2022.

\bibitem{alvarezmelis2018robustness}
D.~Alvarez-Melis and T.~S. Jaakkola, ``On the robustness of interpretability
  methods,'' \emph{arXiv preprint arXiv:1806.08049}, 2018.

\bibitem{10.1007/978-3-030-65474-0_13}
R.~Mangal, K.~Sarangmath, A.~V. Nori, and A.~Orso, ``Probabilistic lipschitz
  analysis of neural networks,'' \emph{International Static Analysis Symposium
  (SAS)}, pp. 274--309, 2020.

\bibitem{arrieta2019explainable}
A.~Barredo~Arrieta, N.~D\'{\i}az-Rodr\'{\i}guez, J.~Del~Ser, A.~Bennetot,
  S.~Tabik, A.~Barbado, S.~Garcia, S.~Gil-Lopez, D.~Molina, R.~Benjamins,
  R.~Chatila, and F.~Herrera, ``Explainable artificial intelligence ({XAI}):
  Concepts, taxonomies, opportunities and challenges toward responsible {AI},''
  \emph{Information Fusion}, vol.~58, p. 82–115, 2020.

\bibitem{lundstrom2022rigorous}
D.~Lundstrom, T.~Huang, and M.~Razaviyayn, ``A rigorous study of integrated
  gradients method and extensions to internal neuron attributions,''
  \emph{Proceedings of the 39th International Conference on Machine Learning},
  vol. 162, pp. 14\,485--14\,508, 2022.

\bibitem{ieeeLip}
T.~Avant and K.~A. Morgansen, ``Analytical bounds on the local lipschitz
  constants of relu networks,'' \emph{IEEE Transactions on Neural Networks and
  Learning Systems}, pp. 1--12, 2023.

\bibitem{gouk2020regularisation}
H.~Gouk, E.~Frank, B.~Pfahringer, and M.~J. Cree, ``Regularisation of neural
  networks by enforcing lipschitz continuity,'' \emph{Machine Learning}, vol.
  110, pp. 393--416, 2020.

\bibitem{NIPS2017_b22b257a}
P.~L. Bartlett, D.~J. Foster, and M.~J. Telgarsky, ``Spectrally-normalized
  margin bounds for neural networks,'' \emph{Advances in Neural Information
  Processing Systems (NeurIPS)}, vol.~30, 2017.

\bibitem{pmlr-v40-Neyshabur15}
B.~Neyshabur, R.~Tomioka, and N.~Srebro, ``Norm-based capacity control in
  neural networks,'' \emph{Proceedings of the 28th Conference on Learning
  Theory}, vol.~40, pp. 1376--1401, 2015.

\bibitem{sanyal2020stable}
A.~Sanyal, P.~Torr, and P.~Dokania, ``Stable rank normalization for improved
  generalization in neural networks and gans,'' \emph{Eighth International
  Conference on Learning Representations (ICLR)}, 2020.

\bibitem{Federer_2005}
H.~Federer, \emph{Geometric measure theory}.\hskip 1em plus 0.5em minus
  0.4em\relax Springer, 2005.

\bibitem{bommer2023finding}
P.~Bommer, M.~Kretschmer, A.~Hedström, D.~Bareeva, and M.~M.~C. Höhne,
  ``Finding the right {XAI} method -- a guide for the evaluation and ranking of
  explainable {AI} methods in climate science,'' \emph{arXiv preprint
  arXiv:2303.00652}, 2023.

\bibitem{wang2020smoothed}
Z.~Wang, H.~Wang, S.~Ramkumar, M.~Fredrikson, P.~Mardziel, and A.~Datta,
  ``Smoothed geometry for robust attribution,'' \emph{Proceedings of the 34th
  International Conference on Neural Information Processing Systems}, 2020.

\end{thebibliography}

\end{document}